%% file: main.tex
\PassOptionsToPackage{table}{xcolor}
\ifdefined\pdfminorversion\pdfminorversion=7\fi
\documentclass[sigconf]{acmart}

\usepackage{graphicx}
\usepackage{booktabs}
\usepackage{array}
\usepackage[table]{xcolor}
\usepackage{multirow}

\makeatletter
\patchcmd{\@mkbibcitation}{\ref{TotPages}}{\pageref{LastMainPaperPage}}{}{}
\patchcmd{\@mkbibcitation}{\ref{TotPages}}{\pageref{LastMainPaperPage}}{}{}
\makeatother

\copyrightyear{2026}
\acmYear{2026}
\setcopyright{cc}
\setcctype{by}
\acmConference[MM '26]{Proceedings of the 34th ACM International Conference on Multimedia}{November 10--14, 2026}{Rio de Janeiro, Brazil}
\acmBooktitle{Proceedings of the 34th ACM International Conference on Multimedia (MM '26), November 10--14, 2026, Rio de Janeiro, Brazil}
\acmDOI{10.1145/3767308.3836337}
\acmISBN{979-8-4007-2213-4/2026/11}

\definecolor{tabbg}{gray}{0.93}
\definecolor{refblue}{rgb}{0.10,0.25,0.95}

\newcolumntype{C}[1]{>{\centering\arraybackslash}m{#1}}
\newcolumntype{L}[1]{>{\raggedright\arraybackslash}m{#1}}

\newcolumntype{S}{!{\hspace{3pt}\vrule width 0.5pt\hspace{3pt}}}

\newcolumntype{V}{!{\hspace{6pt}\vrule width 0.5pt\hspace{6pt}}}
\newcolumntype{Q}{!{\hspace{1pt}\vrule width 0.5pt\hspace{1pt}}}

\newcolumntype{G}{>{\columncolor{tabbg}[0pt][0pt]\centering\arraybackslash}m{4.7em}}
\newcolumntype{Y}{>{\columncolor{tabbg}[0pt][0pt]\centering\arraybackslash}m{4.8em}}
\newcolumntype{Z}{>{\columncolor{tabbg}[0pt][0pt]\centering\arraybackslash}m{4.5em}}

\begin{document}

\title{RiSE-Drive: Risk-Sensitive Long-Horizon Candidate Evaluation for End-to-End Autonomous Driving}

\author{Yu Zhao}
\affiliation{%
  \department{School of Software Technology}
  \institution{Zhejiang University}
  \city{Ningbo}
  \country{China}
}
\email{y.zhao@zju.edu.cn}

\author{Hongyi Duan}
\affiliation{%
  \department{School of Software Technology}
  \institution{Zhejiang University}
  \city{Ningbo}
  \country{China}
}
\email{22551120@zju.edu.cn}

\author{Ming Yin}
\affiliation{%
  \department{School of Software Technology}
  \institution{Zhejiang University}
  \city{Ningbo}
  \country{China}
}
\email{ming.yin@zju.edu.cn}

\author{Hongyuan Zhang}
\affiliation{%
  \department{School of Software Technology}
  \institution{Zhejiang University}
  \city{Ningbo}
  \country{China}
}
\email{metastarxx@gmail.com}

\author{Fan Yang}
\affiliation{%
  \department{School of Software Technology}
  \institution{Zhejiang University}
  \city{Hangzhou}
  \country{China}
}
\email{fyoung@zju.edu.cn}

\author{Hongzhi Wang}
\affiliation{%
  \department{School of Software Technology}
  \institution{Zhejiang University}
  \city{Ningbo}
  \country{China}
}
\email{1030168088@qq.com}

\author{Xiubo Liang}
\affiliation{%
  \department{School of Software Technology}
  \institution{Zhejiang University}
  \city{Ningbo}
  \country{China}
}
\email{xiubo@zju.edu.cn}

\begin{abstract}
  End-to-end autonomous driving systems increasingly rely on generating multiple trajectory candidates and selecting one for execution. In this setting, a central challenge is no longer only candidate generation, but how to evaluate candidates according to their long-horizon consequences under uncertainty. Existing methods still mainly use deterministic point scores, heuristic subscores, or single-outcome evaluations, which makes it difficult to capture \textbf{delayed failures}, \textbf{tail risk}, and ranking instability under plausible future perturbations. We present \textbf{RiSE-Drive}, a framework for \textbf{risk-sensitive long-horizon candidate evaluation} in end-to-end autonomous driving. RiSE-Drive first performs candidate-conditioned latent future rollout, then estimates \textbf{long-horizon outcome distributions} instead of single scalar scores, and finally ranks candidates using a tail-risk-aware and perturbation-robust decision criterion. This reformulates candidate scoring from deterministic oracle-style evaluation into distributional outcome assessment with explicit risk control. As a result, the evaluator is encouraged to prefer candidates that are not only strong under nominal conditions, but also reliable under long-horizon rollout and safety-critical deviations. Experiments on the \textbf{NAVSIM-v1 and NAVSIM-v2 benchmarks}, as well as the \textbf{closed-loop HUGSIM benchmark}, show that RiSE-Drive achieves state-of-the-art performance and significantly improves the robustness of candidate selection.
\end{abstract}



\begin{teaserfigure}
  \includegraphics[width=\textwidth]{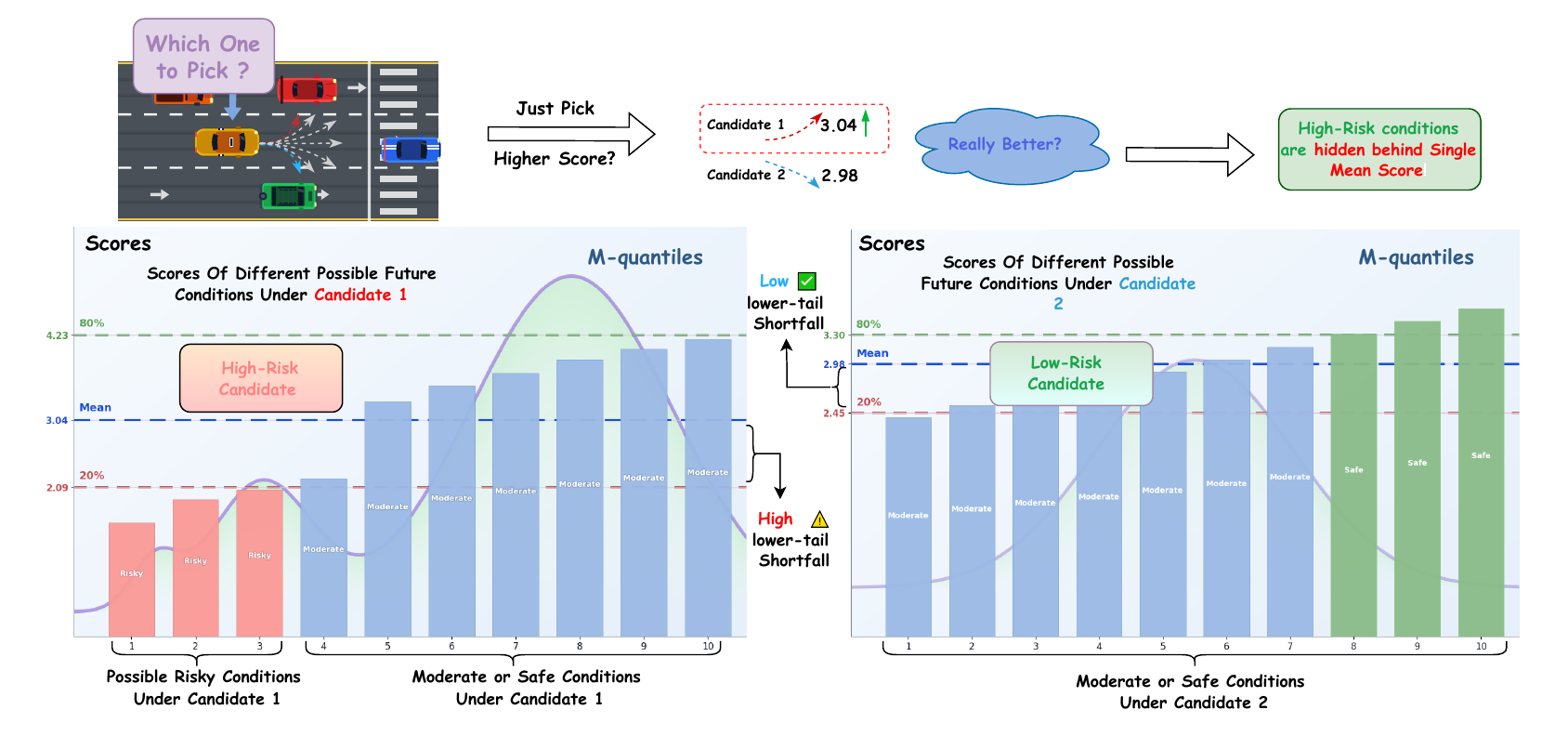}
  \caption{\textbf{Why deterministic mean scoring can fail.} Two ego candidates may have similar mean scores, yet one can still hide a small set of catastrophic future-condition branches. RiSE-Drive therefore evaluates each candidate under multiple future-condition branches, forms a distribution over branch-level long-horizon outcomes, and ranks candidates with risk-sensitive statistics such as the mean, lower quantiles, and CVaR.}
  \Description{RiSE-Drive evaluates each candidate under multiple future-condition branches, rolls each branch forward to obtain one long-horizon outcome sample, forms a distribution over branch outcomes, and ranks candidates using risk-sensitive statistics including the mean, lower quantiles, and CVaR.}
  \label{fig:teaser}
\end{teaserfigure}

\maketitle

\section{Introduction}

End-to-end autonomous driving aims to map raw multi-view observations directly to a safe, rule-compliant, and route-progressive future ego trajectory in a unified learned planner~\cite{ref:VAD,ref:TransFuser,ref:UniAD,ref:PARA-Drive}. Modern systems increasingly adopt a generate-and-select paradigm, where a proposal module enumerates multiple feasible futures and a selector chooses one for execution~\cite{ref:Hydra-MDP,ref:Hydra-MDP++,ref:DrivoR,ref:ipad}. Under behavior-grounded evaluation such as NAVSIM and NAVSIM-v2~\cite{ref:Human-driver,ref:NAVSIM-v2}, the central challenge is therefore not only to generate plausible candidates, but to evaluate which candidate remains best under long-horizon interactions and future uncertainty.

Among recent ResNet-34-based planners, MindDrive~\cite{ref:MindDrive-NavSim} is a strong representative of richer future reasoning. Yet, at the surface level, even strong planners can remain brittle in safety-critical long-tail scenes and perturbed rollouts: several candidates may receive similar nominal scores, while the selected one fails after downstream interactions. The deeper reason is that candidate quality is not a deterministic scalar, but a distribution of long-horizon consequences induced by uncertain future conditions. Existing methods still largely compress this uncertainty into point scores, heuristic subscores, or single-outcome estimates, without explicit candidate-conditioned counterfactual rollout and lower-tail risk modeling. As a result, hidden tail failures remain invisible to the selector, and ranking becomes unstable exactly when robustness matters most.

We address this limitation with \textbf{RiSE-Drive}, a risk-sensitive long-horizon candidate evaluator for end-to-end driving. RiSE-Drive leaves the upstream perception and proposal-generation pipeline unchanged and modifies only the evaluation stage. For each candidate, it predicts a compact set of future-condition branches, performs candidate-conditioned latent consequence propagation, decodes one long-horizon outcome sample for each branch, estimates an outcome distribution through ordered quantiles, and ranks candidates using a criterion that combines the mean, lower-tail Conditional Value-at-Risk (CVaR)~\cite{ref:CVaR}, and cross-condition instability. This design directly targets the root cause above: branchwise rollout captures delayed interaction effects, distributional estimation exposes hidden tail failures, and CVaR-based ranking favors trajectories that remain reliable not only in expectation but also in the lower tail under plausible future deviations. Instantiated on top of DrivoR~\cite{ref:DrivoR}, RiSE-Drive is orthogonal to backbone scaling and isolates the value of better candidate evaluation.

Our contributions are summarized as follows:
\begin{enumerate}
    \item We identify a central evaluation mismatch in candidate-based end-to-end driving: candidate quality should be modeled as a distribution over future-condition-dependent long-horizon outcomes, rather than reduced to a single deterministic score.
    \item We propose \textbf{RiSE-Drive}, a unified evaluator that combines future-condition modeling, candidate-conditioned latent consequence propagation, distributional outcome estimation, and CVaR-based risk-sensitive ranking to make trajectory selection explicitly tail-risk aware.
    \item \textbf{RiSE-Drive} achieves state-of-the-art results on NAVSIM-v1~\cite{ref:Human-driver}, NAVSIM-v2~\cite{ref:NAVSIM-v2}, and HUGSIM~\cite{ref:HUGSIM}. Under a matched DrivoR setup~\cite{ref:DrivoR}, it improves PDMS from 93.1 to 94.2, EPDMS from 48.3 to 51.3, and average HD-Score from 35.7 to 39.5, while consistently improving safety-critical and closed-loop robustness on hard and perturbation-heavy settings.
\end{enumerate}

\section{Related Work}

\textbf{End-to-end autonomous driving planning.}
Viewed chronologically, the development of end-to-end planning is less a story of ever-heavier perception backbones than a redefinition of the planner's output space. Early systems such as VAD, TransFuser, UniAD, and PARA-Drive established the viability of unified planners built on learned scene representations, stronger sensor fusion, and planning-oriented supervision \cite{ref:VAD,ref:TransFuser,ref:UniAD,ref:PARA-Drive}. The next bottleneck, however, was objective mismatch: optimizing a single demonstrated trajectory encourages an average-case imitation policy, even though urban driving is fundamentally set-valued and often admits multiple near-optimal behaviors. This motivated a second wave that recast planning as sequence modeling or density estimation, including autoregressive methods \cite{ref:DrivingGPT,ref:ARTEMIS}, probabilistic or generative planners \cite{ref:GenAD,ref:VADv2,liu2026safegen}, and diffusion/flow-based formulations \cite{ref:GoalFlow,ref:DiffusionDrive,ref:TransDiffuser,ref:MeanFuser,ref:GuideFlow,ref:DIVER}. In this regime, the decisive question is not merely whether a model can output a plausible future, but whether its decision space preserves sufficient behavioral support to remain reliable when multiple futures are simultaneously feasible. NAVSIM and NAVSIM-v2 make this shift explicit, since their behavior-grounded metrics reward safety, progress, compliance, and robustness under perturbation rather than mere open-loop trajectory closeness \cite{ref:Human-driver,ref:NAVSIM-v2,liu2026adversarial,zhao2026nsfhrptneuralsemanticfield}.

\textbf{Candidate generation, trajectory scoring, and selection.}
An increasingly influential design pattern formulates E2E planning as maintaining a candidate set and learning a selector over it. Early structured planners such as SparseDrive already expose this decomposition through sparse scene representations and explicit proposal reasoning \cite{ref:SparseDrive}, while Hydra-MDP and Hydra-MDP++ make multimodal candidate generation a first-class objective via hydra-distillation \cite{ref:Hydra-MDP,ref:Hydra-MDP++}. Under NAVSIM-style evaluation \cite{ref:Human-driver,ref:NAVSIM-v2}, this decomposition becomes even more consequential: the main difficulty is often not producing nominally plausible trajectories, but deciding which candidate is safest and most decision-optimal under safety, progress, and rule-compliance trade-offs. Accordingly, TransDiffuser emphasizes the support quality and decorrelation of the candidate set \cite{ref:TransDiffuser}; WoTE and World4Drive rank candidates through future-aware world modeling \cite{ref:WoTE,ref:World4Drive}; and DriveSuprim, GTRS, ZTRS, and DrivoR make trajectory scoring itself a primary design axis \cite{ref:DriveSuprim,ref:GTRS-DP,ref:ZTRS,ref:DrivoR}. Among them, DrivoR is a particularly relevant baseline: it couples efficient transformer-based perception with disentangled proposal generation and oracle-supervised sub-score prediction, showing that lightweight proposal-and-selection pipelines can be highly competitive on NAVSIM \cite{ref:DrivoR}. Importantly, DrivoR is not a plain point predictor---it produces multiple proposals and learns to rank them---but its proposal set is still generated by a one-shot deterministic decoder from fixed learned queries under winner-takes-all supervision, and the selector operates without explicit action-conditioned rollout of future scene responses, forcing long-horizon consequences to be inferred from the current compressed scene representation alone \cite{ref:DrivoR}. As a result, when several candidates are locally plausible but diverge mainly through downstream interactions, selection quality can still be bottlenecked by limited counterfactual consequence modeling. Related proposal-centric formulations such as ComDrive and iPad further reinforce that modern E2E planning is increasingly governed by candidate coverage, scoring fidelity, and risk-sensitive selection rather than raw trajectory regression \cite{ref:ComDrive,ref:ipad}.

\section{Preliminary}

\subsection{Candidate-Based End-to-End Driving with DrivoR}
Candidate-based end-to-end driving methods first generate a finite set of feasible future ego trajectories and then select one trajectory for execution via a learned evaluation module. DrivoR provides a representative generate-and-score formulation and serves as the baseline planning pipeline in our framework. It consists of a compact multi-camera scene encoder, a proposal generator that outputs multiple trajectory candidates, and a scoring module that evaluates these candidates by predicted driving-quality sub-scores.

At discrete time step $t$, the observation is defined as
\begin{equation}
o_t = \left( \left\{ \mathbf{I}_t^{(n)} \right\}_{n=1}^{N}, \mathbf{e}_t \right),
\end{equation}
where $\mathbf{I}_t^{(n)} \in \mathbb{R}^{H_I \times W_I \times 3}$ is the RGB image from the $n$-th onboard camera, $N$ is the number of cameras, and $\mathbf{e}_t \in \mathbb{R}^{d_e}$ is the ego-state vector containing pose, velocity, acceleration, and navigation command.

The perception encoder maps the observation to a compact scene-token representation
\begin{equation}
\mathbf{R}_t = F_{\mathrm{enc}}(o_t) \in \mathbb{R}^{N_r \times D},
\end{equation}
where $N_r$ is the number of scene tokens and $D$ is the token dimension. The proposal generator then predicts a set of $K$ candidate ego trajectories,
\begin{equation}
\mathcal{T}_t
=
\left\{ \tau_1, \tau_2, \ldots, \tau_K \right\}
=
G_{\mathrm{prop}}(\mathbf{R}_t, \mathbf{e}_t).
\end{equation}
Each candidate trajectory is a sequence of $H$ future ego states,
\begin{equation}
\tau_i
=
\left\{ \mathbf{p}_{i,h} \right\}_{h=1}^{H},
\qquad
\mathbf{p}_{i,h} = (x_{i,h}, y_{i,h}, \theta_{i,h}) \in \mathbb{R}^{3},
\end{equation}
where $x_{i,h}$ and $y_{i,h}$ denote the longitudinal and lateral displacements in the ego-local frame, and $\theta_{i,h}$ denotes the heading angle at prediction step $h$.

Given the candidate set $\mathcal{T}_t$, the scoring network predicts a $C$-dimensional sub-score vector for each candidate,
\begin{equation}
\mathbf{s}_i
=
G_{\mathrm{score}}(\mathbf{R}_t, \tau_i)
\in \mathbb{R}^{C},
\end{equation}
and aggregates it into a scalar score
\begin{equation}
u_i = \boldsymbol{\lambda}^{\top}\mathbf{s}_i,
\qquad
\boldsymbol{\lambda} \in \mathbb{R}^{C}.
\end{equation}
The final plan is selected as
\begin{equation}
\tau_t^{\star}
=
\arg\max_{\tau_i \in \mathcal{T}_t} u_i.
\end{equation}

This formulation separates candidate generation from candidate evaluation: the proposal module produces diverse feasible futures, and the scoring module selects the candidate to execute.

\subsection{From Deterministic Candidate Scores to Long-Horizon Outcome Variables}
The deterministic scoring rule above can be written compactly as
\begin{equation}
u_i = f_{\psi}(o_t, \tau_i),
\end{equation}
where $f_{\psi}$ denotes the candidate scoring function induced by the perception encoder, the scoring module, and the final linear aggregation, and $\psi$ denotes its parameters.

The quality of a candidate is determined by the long-horizon consequences induced by executing that candidate. Let $\omega$ denote a future-condition variable and define
\begin{equation}
Y_i = Y(\tau_i, \omega),
\qquad
\omega \sim p(\omega \mid o_t),
\end{equation}
where $Y_i \in \mathbb{R}$ is the long-horizon outcome associated with candidate $\tau_i$, and $p(\omega \mid o_t)$ is the conditional distribution of future conditions given the current observation. Under this formulation, the deterministic score $u_i$ is only a point estimate of $Y_i$ and does not model the distribution of future consequences.

Our objective is therefore to replace pointwise candidate evaluation with distributional candidate evaluation. Instead of ranking candidates by a single deterministic score, we rank them using statistics of the long-horizon outcome distribution, including its mean, lower-tail risk, and cross-condition stability.

\section{Method}

\subsection{Overview}

\begin{figure*}[t]
    \centering
    \includegraphics[width=\textwidth]{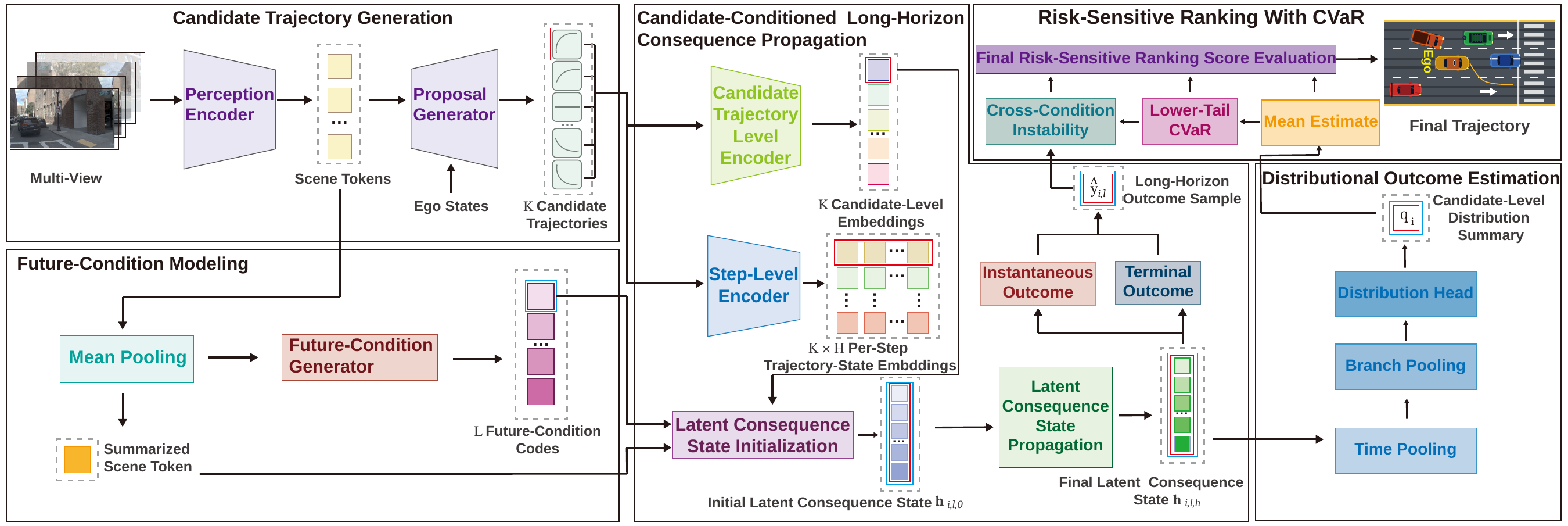}
    \caption{\textbf{Overview of the RiSE-Drive pipeline.} Multi-view observations are encoded, and the proposal generator produces $K$ ego-trajectory candidates. A future-condition generator predicts $L$ branch-specific condition codes that parameterize plausible downstream evolutions. For each candidate--condition pair, RiSE-Drive initializes a latent consequence state, propagates it along the trajectory, and decodes one branch-level long-horizon outcome sample. The resulting branchwise consequence traces are pooled over time and across branches and mapped to ordered outcome quantiles by a distribution head. The final ranking score combines the mean estimate, lower-tail CVaR, and cross-condition instability, thereby favoring candidates that remain reliable under plausible future deviations rather than merely optimal under a single nominal forecast.}
    \Description{Overview of RiSE-Drive. Multi-view observations are encoded into scene features. The proposal generator creates candidate ego trajectories, while the future-condition generator creates plausible downstream branches. Each candidate and branch pair is propagated to obtain a long-horizon outcome sample. The resulting distribution is summarized with ordered quantiles and ranked using the mean, lower-tail CVaR, and cross-condition instability.}
    \label{fig:rise_pipeline}
\end{figure*}

RiSE-Drive leaves the upstream perception and proposal generation pipeline unchanged and modifies only the candidate-evaluation stage. As summarized in Fig.~\ref{fig:rise_pipeline}, it predicts a small set of future-condition branches for each scene, rolls each candidate under each branch to obtain branch-level long-horizon outcome samples, aggregates these samples into a candidate-level outcome distribution, and finally ranks candidates with a risk-sensitive criterion based on the mean, lower-tail CVaR, and cross-condition instability.

\subsection{Future-Condition Modeling}
We first summarize the scene tokens by mean pooling:
\begin{equation}
\bar{\mathbf{r}}_t
=
\frac{1}{N_r}\sum_{j=1}^{N_r}\mathbf{R}_t^{(j)}
\in \mathbb{R}^{D},
\end{equation}
where $\mathbf{R}_t^{(j)} \in \mathbb{R}^{D}$ is the $j$-th scene token. Based on $\bar{\mathbf{r}}_t$, the future-condition generator produces $L$ branch-specific future-condition codes,
\begin{equation}
\omega_l = P_{\mathrm{ctx}}^{(l)}(\bar{\mathbf{r}}_t) \in \mathbb{R}^{d_{\omega}},
\qquad
l=1,\ldots,L.
\end{equation}
The resulting set
\begin{equation}
\Omega_t = \left\{ \omega_1, \ldots, \omega_L \right\}
\end{equation}
defines the set of rollout branches used in candidate evaluation. Each $\omega_l$ is a learned condition code used to initialize and modulate one rollout branch.

\subsection{Candidate-Conditioned Long-Horizon Consequence Propagation}
For each candidate trajectory $\tau_i$, we compute a candidate-level embedding
\begin{equation}
\mathbf{c}_i = E_{\tau}(\tau_i) \in \mathbb{R}^{d_c},
\end{equation}
where $E_{\tau}$ is the \emph{candidate trajectory level encoder}. We further define a step-wise trajectory-state embedding
\begin{equation}
\mathbf{z}_{i,h} = E_p(\mathbf{p}_{i,h}) \in \mathbb{R}^{d_p},
\qquad
h=1,\ldots,H,
\end{equation}
where $E_p$ is the \emph{step level encoder}.

For each candidate--condition pair $(\tau_i,\omega_l)$, the initial latent consequence state is
\begin{equation}
\mathbf{h}_{i,l,0}
=
W_r \bar{\mathbf{r}}_t
+
W_c \mathbf{c}_i
+
W_{\omega} \omega_l
+
\mathbf{b}_0,
\end{equation}
where $W_r \in \mathbb{R}^{d_h \times D}$, $W_c \in \mathbb{R}^{d_h \times d_c}$, $W_{\omega} \in \mathbb{R}^{d_h \times d_{\omega}}$, and $\mathbf{b}_0 \in \mathbb{R}^{d_h}$ are learnable parameters.

We then propagate the latent consequence state over the evaluation horizon according to
\begin{equation}
\mathbf{h}_{i,l,h}
=
T_{\eta}(\mathbf{h}_{i,l,h-1}, \mathbf{z}_{i,h}, \bar{\mathbf{r}}_t),
\qquad
h=1,\ldots,H,
\end{equation}
with
\begin{equation}
\tilde{\mathbf{h}}_{i,l,h}
=
\phi\!\left(
W_h \mathbf{h}_{i,l,h-1}
+
W_p \mathbf{z}_{i,h}
+
W_s \bar{\mathbf{r}}_t
+
\mathbf{b}_h
\right),
\end{equation}
\begin{equation}
\mathbf{g}_{i,l,h}
=
\sigma\!\left(
U_h \mathbf{h}_{i,l,h-1}
+
U_p \mathbf{z}_{i,h}
+
U_s \bar{\mathbf{r}}_t
+
\mathbf{b}_g
\right),
\end{equation}
\begin{equation}
\mathbf{h}_{i,l,h}
=
\mathbf{g}_{i,l,h}\odot \tilde{\mathbf{h}}_{i,l,h}
+
(1-\mathbf{g}_{i,l,h})\odot \mathbf{h}_{i,l,h-1},
\end{equation}
where $\phi(\cdot)$ is the GELU nonlinearity, $\sigma(\cdot)$ is the sigmoid function, and $\odot$ denotes element-wise multiplication.

At each step, we decode an instantaneous outcome
\begin{equation}
r_{i,l,h} = g_r(\mathbf{h}_{i,l,h}) \in \mathbb{R},
\end{equation}
and from the final state we decode a terminal outcome
\begin{equation}
r_{i,l}^{\mathrm{term}} = g_T(\mathbf{h}_{i,l,H}) \in \mathbb{R}.
\end{equation}
The long-horizon outcome sample under condition $\omega_l$ is defined as
\begin{equation}
\hat{y}_{i,l}
=
\sum_{h=1}^{H}\gamma^{h-1} r_{i,l,h}
+
\gamma^{H} r_{i,l}^{\mathrm{term}}.
\end{equation}
For a fixed candidate $\tau_i$, the set
\begin{equation}
\left\{ \hat{y}_{i,l} \right\}_{l=1}^{L}
\end{equation}
provides a finite-sample representation of its long-horizon outcome distribution.

\subsection{Distributional Outcome Estimation}
We first pool each branch-specific latent consequence trajectory over time,
\begin{equation}
\bar{\mathbf{h}}_{i,l}
=
\frac{1}{H}\sum_{h=1}^{H}\mathbf{h}_{i,l,h}
\in \mathbb{R}^{d_h},
\end{equation}
and then pool across branches,
\begin{equation}
\tilde{\mathbf{h}}_i
=
\frac{1}{L}\sum_{l=1}^{L}\bar{\mathbf{h}}_{i,l}
\in \mathbb{R}^{d_h}.
\end{equation}

Based on $\tilde{\mathbf{h}}_i$, the distribution head outputs $M$ ordered quantiles,
\begin{equation}
\mathbf{q}_i
=
\left[
q_i^{(1)},
q_i^{(2)},
\ldots,
q_i^{(M)}
\right]^{\top}
\in \mathbb{R}^{M},
\end{equation}
where $q_i^{(m)}$ corresponds to quantile level $\alpha_m$ and
\begin{equation}
0 < \alpha_1 < \alpha_2 < \cdots < \alpha_M < 1.
\end{equation}
To enforce monotonicity, we use
\begin{equation}
q_i^{(1)} = a_i,
\qquad
a_i = h_0(\tilde{\mathbf{h}}_i),
\end{equation}
\begin{equation}
q_i^{(m)}
=
q_i^{(m-1)} + \mathrm{softplus}\!\left(h_m(\tilde{\mathbf{h}}_i)\right),
\qquad
m=2,\ldots,M,
\end{equation}
which guarantees
\begin{equation}
q_i^{(1)} \le q_i^{(2)} \le \cdots \le q_i^{(M)}.
\end{equation}
We define the mean estimate of the long-horizon outcome distribution as
\begin{equation}
\hat{\mu}_i
=
\frac{1}{M}\sum_{m=1}^{M} q_i^{(m)}.
\end{equation}

\subsection{Risk-Sensitive Ranking with CVaR}
A ranking rule based only on $\hat{\mu}_i$ is insensitive to lower-tail failures. To account for tail risk, we use the lower-tail Conditional Value-at-Risk (CVaR). Let
\begin{equation}
M_{\alpha} = \lceil \alpha M \rceil,
\end{equation}
where $\alpha \in (0,1)$ is the lower-tail probability level. The discrete lower-tail CVaR estimate of candidate $\tau_i$ is
\begin{equation}
\widehat{\mathrm{CVaR}}_{\alpha}^{\downarrow}(Y_i)
=
\frac{1}{M_{\alpha}}
\sum_{m=1}^{M_{\alpha}} q_i^{(m)}.
\end{equation}
We further define the lower-tail shortfall
\begin{equation}
\hat{\Delta}_i
=
\hat{\mu}_i
-
\widehat{\mathrm{CVaR}}_{\alpha}^{\downarrow}(Y_i).
\end{equation}

To measure cross-condition instability, we compute the branch-wise standard deviation
\begin{equation}
\bar{y}_i
=
\frac{1}{L}\sum_{l=1}^{L}\hat{y}_{i,l},
\end{equation}
\begin{equation}
\hat{\sigma}_i
=
\sqrt{
\frac{1}{L}\sum_{l=1}^{L}
\left(
\hat{y}_{i,l} - \bar{y}_i
\right)^2
}.
\end{equation}

The final risk-sensitive ranking score is defined as
\begin{equation}
\hat{J}_i
=
\hat{\mu}_i
-
\lambda_{\mathrm{tail}} \hat{\Delta}_i
-
\lambda_{\mathrm{var}} \hat{\sigma}_i,
\end{equation}
where $\lambda_{\mathrm{tail}} \ge 0$ and $\lambda_{\mathrm{var}} \ge 0$ are trade-off coefficients. Candidates with larger $\hat{J}_i$ are preferred.

\subsection{Training Objectives}
For each candidate--condition pair $(\tau_i,\omega_l)$, we assume a supervision signal
\begin{equation}
y_{i,l}^{\star} \in \mathbb{R},
\qquad
i=1,\ldots,K,\;\; l=1,\ldots,L,
\end{equation}
which represents the target long-horizon outcome under the corresponding future-condition branch.

We first supervise branch-level long-horizon outcome prediction using the Huber loss
\begin{equation}
\mathcal{L}_{\mathrm{branch}}
=
\frac{1}{KL}
\sum_{i=1}^{K}\sum_{l=1}^{L}
\ell_{\mathrm{Huber}}(\hat{y}_{i,l}, y_{i,l}^{\star}).
\end{equation}

We then supervise the predicted quantiles using the pinball loss
\begin{equation}
\rho_{\alpha}(u)
=
u\left(\alpha - \mathbf{1}[u<0]\right),
\end{equation}
\begin{equation}
\mathcal{L}_{\mathrm{quant}}
=
\frac{1}{KML}
\sum_{i=1}^{K}\sum_{m=1}^{M}\sum_{l=1}^{L}
\rho_{\alpha_m}\!\left(y_{i,l}^{\star} - q_i^{(m)}\right).
\end{equation}

We further compute the empirical target statistics
\begin{equation}
\mu_i^{\star}
=
\frac{1}{L}\sum_{l=1}^{L} y_{i,l}^{\star},
\end{equation}
\begin{equation}
\sigma_i^{\star}
=
\sqrt{
\frac{1}{L}\sum_{l=1}^{L}
\left(
y_{i,l}^{\star} - \mu_i^{\star}
\right)^2
},
\end{equation}
and the empirical lower-tail CVaR
\begin{equation}
\mathrm{CVaR}_{\alpha}^{\star \downarrow}(Y_i).
\end{equation}
The corresponding target ranking score is
\begin{equation}
J_i^{\star}
=
\mu_i^{\star}
-
\lambda_{\mathrm{tail}}
\left(
\mu_i^{\star} - \mathrm{CVaR}_{\alpha}^{\star \downarrow}(Y_i)
\right)
-
\lambda_{\mathrm{var}} \sigma_i^{\star}.
\end{equation}

We optimize candidate ordering with a pairwise ranking loss
\begin{equation}
\mathcal{L}_{\mathrm{rank}}
=
\sum_{i=1}^{K}\sum_{\substack{j=1 \\ j \neq i}}^{K}
\max\left(
0,\,
\delta
-
\mathrm{sign}(J_i^{\star}-J_j^{\star})
(\hat{J}_i-\hat{J}_j)
\right),
\end{equation}
where $\delta > 0$ is a margin hyperparameter.

The final objective is
\begin{equation}
\mathcal{L}
=
\lambda_b \mathcal{L}_{\mathrm{branch}}
+
\lambda_q \mathcal{L}_{\mathrm{quant}}
+
\lambda_r \mathcal{L}_{\mathrm{rank}},
\end{equation}
where $\lambda_b$, $\lambda_q$, and $\lambda_r$ are nonnegative loss weights.

\begin{table*}[t]
\centering
\small
\setlength{\tabcolsep}{0pt}
\renewcommand{\arraystretch}{0.90}

\begin{tabular}{@{}L{8.8em}VC{3.6em}VC{7.4em}VC{7.0em}V*{5}{C{3.8em}}QG@{}}
\specialrule{1.1pt}{0pt}{0pt}
Method & Input & Backbone & Venue & NC$\uparrow$ & DAC$\uparrow$ & EP$\uparrow$ & TTC$\uparrow$ & Conf.$\uparrow$ & PDMS$\uparrow$ \\
\specialrule{0.7pt}{0.15em}{0.18em}

Human & - & - & - & 100 & 100 & 87.5 & 100 & 99.9 & 94.8 \\

\specialrule{0.7pt}{0.18em}{0.18em}

DrivingGPT \cite{ref:DrivingGPT} & C & VQ-VAE/GPT & ICCV 2025 & 98.9 & 90.7 & 79.7 & 94.9 & 95.6 & 82.4 \\
VADv2 \cite{ref:VADv2} & C & ResNet-34 & arXiv 2024 & 97.9 & 91.7 & 77.6 & 92.9 & \textbf{100} & 83.0 \\
UniAD \cite{ref:UniAD} & C & ResNet-34 & CVPR 2023 & 97.8 & 91.9 & 78.8 & 92.9 & \textbf{100} & 83.4 \\
LTF \cite{ref:TransFuser} & C & ResNet-34 & TPAMI 2022 & 97.4 & 92.8 & 79.0 & 92.4 & \textbf{100} & 83.8 \\
TransFuser \cite{ref:TransFuser} & C \& L & ResNet-34 & TPAMI 2022 & 97.7 & 92.8 & 79.2 & 92.8 & \textbf{100} & 84.0 \\
PARA-Drive \cite{ref:PARA-Drive} & C & ResNet-34 & CVPR 2024 & 97.9 & 92.4 & 79.3 & 93.0 & 99.8 & 84.0 \\
DriveX-S \cite{ref:DriveX-S} & C & InternImage-Tiny & ICCV 2025 & 97.5 & 94.0 & 79.7 & 93.0 & \textbf{100} & 84.5 \\
LAW \cite{ref:LAW} & C & ResNet-34 & ICLR 2025 & 96.4 & 95.4 & 81.7 & 88.7 & 99.9 & 84.6 \\
World4Drive \cite{ref:World4Drive} & C & ResNet-34 & ICCV 2025 & 97.4 & 94.3 & 79.9 & 92.8 & \textbf{100} & 85.1 \\
DRAMA \cite{ref:DRAMA} & C \& L & ResNet-34 & arXiv 2024 & 98.0 & 93.1 & 80.1 & 94.8 & \textbf{100} & 85.5 \\
GoalFlow \cite{ref:GoalFlow} & C \& L & ResNet-34 & CVPR 2025 & 98.3 & 93.8 & 79.8 & 94.3 & \textbf{100} & 85.7 \\
Hydra-MDP \cite{ref:Hydra-MDP} & C \& L & ResNet-34 & arXiv 2024 & 98.3 & 96.0 & 78.7 & 94.6 & \textbf{100} & 86.5 \\
Hydra-MDP++ \cite{ref:Hydra-MDP++} & C \& L & ResNet-34 & arXiv 2024 & 97.6 & 96.0 & 80.4 & 93.1 & \textbf{100} & 86.6 \\
ARTEMIS \cite{ref:ARTEMIS} & C \& L & ResNet-34 & arXiv 2025 & 98.3 & 95.1 & 81.4 & 94.3 & \textbf{100} & 87.0 \\
DiffusionDrive \cite{ref:DiffusionDrive} & C \& L & ResNet-34 & CVPR 2025 & 98.2 & 96.2 & 82.2 & 94.7 & \textbf{100} & 88.1 \\
WoTE \cite{ref:WoTE} & C \& L & ResNet-34 & ICCV 2025 & 98.5 & 96.8 & 81.9 & 94.9 & 99.9 & 88.3 \\
DIVER \cite{ref:DIVER} & C \& L & ResNet-34 & arXiv 2025 & 98.5 & 96.5 & 82.6 & 94.9 & \textbf{100} & 88.3 \\
MindDrive \cite{ref:MindDrive-NavSim} & C \& L & ResNet-34 & arXiv 2025 & 98.4 & 97.0 & 82.8 & 95.1 & 99.9 & 88.9 \\
MeanFuser \cite{ref:MeanFuser} & C & ResNet-34 & arXiv 2026 & 98.6 & 97.0 & 82.8 & 95.0 & \textbf{100} & 89.0 \\
AutoVLA \cite{ref:AutoVLA} & C & Qwen2.5-VL-3B & NeurIPS 2025 & 98.4 & 95.6 & 81.9 & 98.0 & 99.9 & 89.1 \\
DriveSuprim \cite{ref:DriveSuprim} & C & ResNet-34 & arXiv 2025 & 97.8 & 97.3 & 86.7 & 93.6 & \textbf{100} & 89.9 \\
DriveVLA-W0 \cite{ref:DriveVLA-w0} & C & Emu3-8B & arXiv 2025 & 98.7 & \textbf{99.1} & 83.3 & 95.3 & 99.3 & 90.2 \\
ReCogDrive \cite{ref:ReCogDrive} & C & InternVL3-2B & ICLR 2026 & 97.9 & 97.3 & 87.3 & 94.9 & \textbf{100} & 90.8 \\
iPad \cite{ref:ipad} & C & ResNet-34 & arXiv 2025 & 98.6 & 98.3 & 88.0 & 94.9 & \textbf{100} & 91.7 \\
Centaur \cite{ref:Centaur} & C & V2-99 & arXiv 2025 & \textbf{99.5} & 98.9 & 85.9 & 98.0 & \textbf{100} & 92.6 \\
DrivoR \cite{ref:DrivoR} & C & ViT-S & CVPR 2026 & 98.9 & 98.3 & \textbf{89.1} & 96.2 & \textbf{100} & 93.1 \\

\specialrule{0.7pt}{0.18em}{0.18em}

\textbf{RiSE-Drive (Ours)} & C & ViT-S & - & \textbf{99.5} & 98.5 & 88.7 & \textbf{98.4} & \textbf{100} & \textbf{94.2} \\

\specialrule{1.1pt}{0pt}{0pt}
\end{tabular}

\caption{NAVSIM-v1 evaluation results on the Navtest split. C denotes camera-only input and L denotes LiDAR.}
\label{tab:navsim-v1}
\end{table*}

\begin{table*}[t]
\centering
\small
\setlength{\tabcolsep}{0.8pt}
\renewcommand{\arraystretch}{1.02}

\begin{tabular}{@{}L{8.41em}!{\vrule width 0.4pt}C{5.48em}!{\vrule width 0.55pt}*{9}{C{2.28em}}!{\vrule width 0.55pt}*{9}{C{2.28em}}!{\vrule width 0.55pt}C{3.36em}@{}}
\specialrule{1.1pt}{0pt}{0pt}
& & \multicolumn{9}{c}{Stage 1} & \multicolumn{9}{c}{Stage 2} & \\
Method & Backbone & NC & DAC & DDC & TLC & EP & TTC & LK & HC & EC & NC & DAC & DDC & TLC & EP & TTC & LK & HC & EC & EPDMS \\
\specialrule{0.7pt}{0.20em}{0.22em}

LTF \cite{ref:TransFuser} & ResNet-34
& 96.2 & 79.5 & 99.1 & 99.5 & 84.1 & 95.1 & 94.2 & 97.5 & 79.1
& 77.7 & 70.2 & 84.2 & 98.0 & 85.1 & 75.6 & 45.4 & 95.7 & 75.9
& 23.1 \\

DiffusionDrive$^{*}$ \cite{ref:DiffusionDrive} & ResNet-34
& 96.0 & 79.7 & 97.4 & 99.5 & 81.3 & 93.1 & 90.8 & 96.8 & 73.8
& 82.1 & 72.2 & 88.5 & 98.7 & 85.1 & 78.8 & 49.2 & 89.3 & 71.2
& 24.2 \\

GuideFlow \cite{ref:GuideFlow} & ResNet-34
& 96.6 & 80.5 & 96.3 & 99.3 & 82.3 & 94.9 & 91.5 & \textbf{97.7} & 67.8
& 87.3 & 76.7 & 88.8 & \textbf{99.2} & 84.3 & 85.1 & 49.7 & 93.1 & 44.5
& 27.1 \\

RAP-DINO \cite{ref:RAP-DINO} & ViT-H
& 97.1 & 94.4 & 98.8 & 99.8 & 83.9 & 96.9 & 94.7 & 96.4 & 66.2
& 83.2 & 83.9 & 87.4 & 98.0 & 86.9 & 80.4 & 52.3 & 95.2 & 52.4
& 39.6 \\

DriveSuprim \cite{ref:DriveSuprim} & V2-99
& 98.9 & 95.1 & 99.2 & 99.6 & 76.1 & \textbf{99.1} & 94.7 & 97.6 & 54.2
& 87.9 & 88.8 & 89.6 & 98.8 & 80.3 & 86.0 & 53.5 & 97.1 & 56.1
& 42.1 \\

GTRS-Dense \cite{ref:GTRS-DP} & V2-99
& 98.9 & 96.2 & 99.4 & 99.3 & 72.9 & 98.9 & 95.1 & 96.9 & 39.1
& 91.2 & 89.4 & 94.4 & 98.8 & 69.5 & 90.0 & 54.3 & 94.0 & 48.7
& 45.0 \\

DiffVLA \cite{ref:diffvla} & Vicuna-v1.5
& 95.7 & \textbf{99.2} & \textbf{100} & \textbf{100} & \textbf{85.9} & 96.4 & \textbf{97.1} & 95.0 & \textbf{84.2}
& 81.2 & 88.8 & 94.6 & 99.0 & 86.0 & 76.4 & 59.8 & \textbf{98.6} & \textbf{80.4}
& 45.0 \\

GTRS-Aug \cite{ref:GTRS-DP} & V2-99
& 98.9 & 95.1 & 99.1 & 99.6 & 76.2 & \textbf{99.1} & 94.9 & 97.6 & 54.2
& 88.1 & 88.8 & 89.3 & 98.9 & \textbf{98.9} & 85.9 & 53.7 & 96.8 & 56.9
& 45.4 \\

ZTRS \cite{ref:ZTRS} & V2-99
& 98.9 & 97.6 & \textbf{100} & \textbf{100} & 66.7 & 98.9 & 96.2 & 96.7 & 44.0
& 91.1 & \textbf{90.4} & \textbf{95.8} & 99.0 & 63.6 & 89.8 & \textbf{60.4} & 97.6 & 66.1
& 48.1 \\

DrivoR \cite{ref:DrivoR} & ViT-S
& 98.8 & 95.1 & 98.9 & \textbf{100} & 72.6 & 98.7 & 94.0 & 97.6 & 73.3
& 90.2 & 88.4 & 91.9 & 98.6 & 70.0 & 88.0 & 50.1 & 98.5 & 76.2
& 48.3 \\

\specialrule{0.7pt}{0.18em}{0.18em}
\textbf{RiSE-Drive (Ours)} & ViT-S
& \textbf{99.0} & 96.8 & 99.3 & \textbf{100} & 72.8 & 99.0 & 94.8 & \textbf{97.7} & 73.5
& \textbf{91.5} & 89.8 & 94.6 & 98.8 & 68.7 & \textbf{91.3} & 54.9 & 98.4 & 76.7
& \textbf{51.3} \\

\specialrule{1.1pt}{0pt}{0pt}
\end{tabular}

\caption{NAVSIM-v2 \texttt{navhard-two-stage} results. $^{*}$ denotes results reproduced with the official code repository or official checkpoint.}
\label{tab:navsim-v2-navhard-two-stage}
\end{table*}

\begin{table}[t]
\centering
\small
\setlength{\tabcolsep}{0pt}
\renewcommand{\arraystretch}{0.95}

\begin{tabular}{@{}L{4.75em}S*{5}{C{2.35em}}S*{5}{C{2.35em}}@{}}
\specialrule{1.1pt}{0pt}{0pt}
Method & \multicolumn{5}{c}{RC} & \multicolumn{5}{c}{HD-Score} \\
       & E & M & H & X & Avg. & E & M & H & X & Avg. \\
\specialrule{0.7pt}{0.18em}{0.18em}

VAD \cite{ref:VAD}
& 51.3          & 31.1          & 25.3          & 26.5          & 31.4
& 36.3          & 9.5           & 8.0           & 11.5          & 13.4 \\

LTF \cite{ref:TransFuser}
& 67.8          & 35.1          & 26.2          & 40.5          & 38.9
& 58.9          & 18.0          & 9.8           & 25.9          & 23.7 \\

UniAD \cite{ref:UniAD}
& 78.4                & \textbf{60.5} & 33.6                & 17.8                & 45.9
& 64.9                & \textbf{45.8} & 20.6                & 6.6                 & 32.7 \\

DrivoR \cite{ref:DrivoR}
& \textbf{80.9} & 50.5          & 33.8          & \textbf{47.1} & 49.8
& 73.3          & 34.6          & 18.8          & 32.5          & 35.7 \\

\specialrule{0.7pt}{0.18em}{0.18em}

\textbf{Ours}
& 76.5          & 53.3          & \textbf{34.1} & 46.3          & \textbf{50.1}
& \textbf{74.6} & 37.2          & \textbf{22.7} & \textbf{39.2} & \textbf{39.5} \\

\specialrule{1.1pt}{0pt}{0pt}
\end{tabular}

\vspace{0.2em}
\begin{minipage}{\linewidth}
\footnotesize \textit{E}: Easy,\ \textit{M}: Medium,\ \textit{H}: Hard,\ \textit{X}: Extreme.
\end{minipage}

\caption{Photorealistic closed-loop evaluation on HUGSIM.}
\label{tab:rc-hdscore-breakdown}
\end{table}

\begin{figure*}[t]
    \centering
     \includegraphics[width=\textwidth]{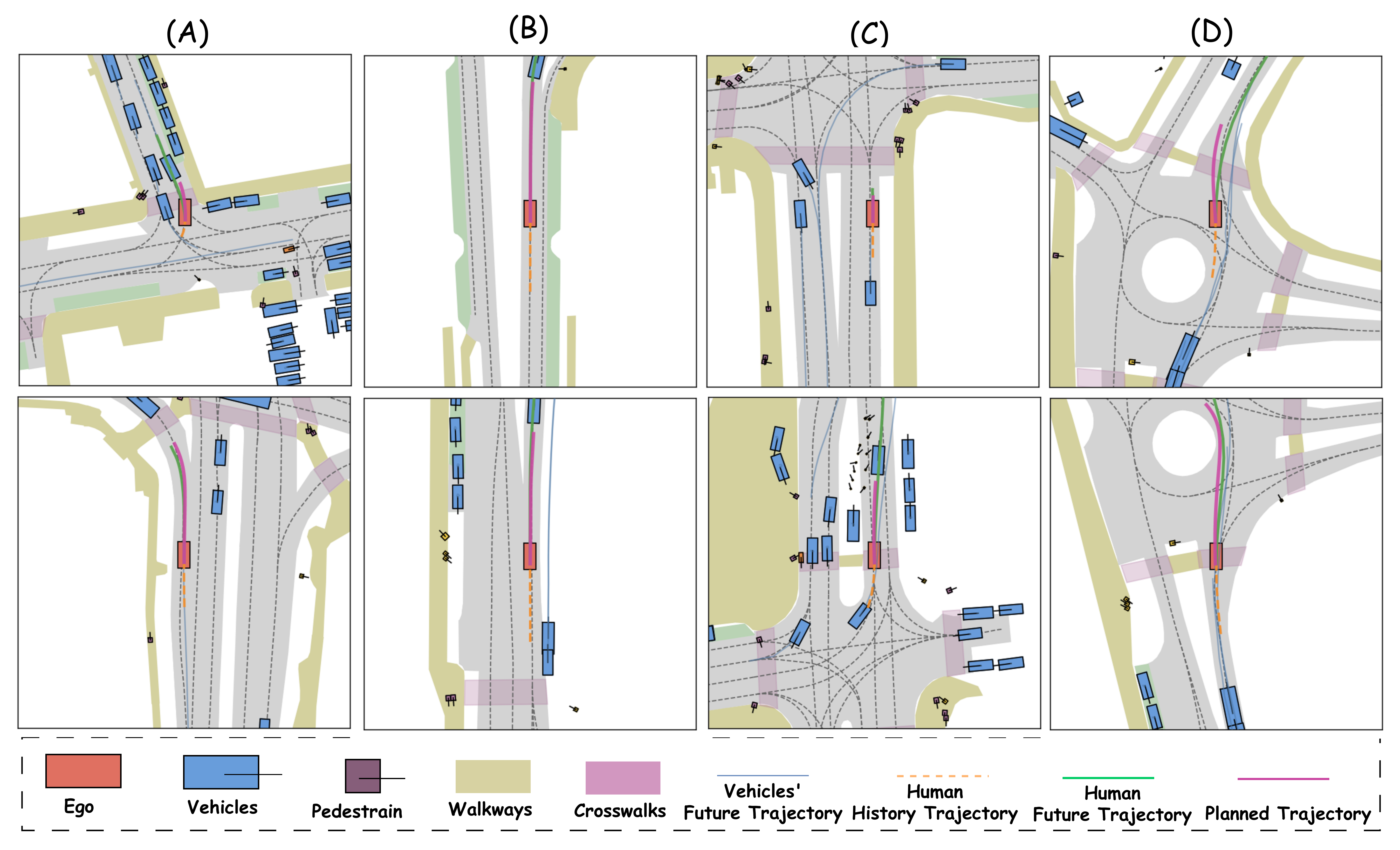}
    \caption{\textbf{Qualitative results of RiSE-Drive on representative NavSafe safety-critical scenarios.} From left to right, columns (a)--(d) correspond to \textbf{UNPL} (\emph{unprotected left turn}), \textbf{EXR} (\emph{exit ramp}), \textbf{YLD} (\emph{yielding}), and \textbf{RDBT} (\emph{roundabout}). Each column shows two temporal snapshots of one representative case. Magenta denotes the RiSE-Drive plan, green the human future trajectory, orange dashed the ego history, and blue the future trajectories of surrounding vehicles. Across all four cases, RiSE-Drive produces smooth, route-consistent, and interaction-aware behaviors while remaining close to safe human driving in conflict-prone regions.}
    \Description{Four columns show unprotected-left-turn, exit-ramp, yielding, and roundabout scenarios, with two temporal snapshots in each column. Magenta curves are RiSE-Drive plans, green curves are human future trajectories, orange dashed curves are ego histories, and blue curves are future trajectories of surrounding vehicles.}
    \label{fig:rise_navsafe_vis}
\end{figure*}

\begin{table}[t]
\centering
\small
\setlength{\tabcolsep}{4.0pt}
\renewcommand{\arraystretch}{1.06}
\resizebox{\columnwidth}{!}{%
\begin{tabular}{l|cc|cc}
\toprule
& \multicolumn{2}{c|}{NAVSIM-v1} & \multicolumn{2}{c}{NAVSIM-v2 NavHard} \\
Configuration & PDMS$\uparrow$ & NC$\uparrow$ & EPDMS$\uparrow$ & TTC$\uparrow$ \\
\midrule
B-I: DrivoR & 93.1 & 98.9 & 48.3 & 88.0 \\
B-II: DrivoR w/o SD & 91.9 & 98.4 & 45.6 & 84.6 \\
\midrule
+ FCG & 92.6 & 98.7 & 46.8 & 86.5 \\
+ CTE & 92.9 & 98.8 & 47.5 & 87.3 \\
+ SLE + LCP & 93.5 & 99.1 & 49.2 & 89.6 \\
+ DOE & 93.9 & 99.3 & 50.6 & 90.5 \\
+ RSR & \textbf{94.2} & \textbf{99.5} & \textbf{51.3} & \textbf{91.3} \\
\bottomrule
\end{tabular}}
\caption{Component-wise ablation of the proposed evaluator. Starting from the reduced baseline B-II, each row adds one RiSE-Drive component to the configuration immediately above.}
\label{tab:ablation_main_single}
\end{table}

\begin{table}[t]
\centering
\small
\setlength{\tabcolsep}{4.2pt}
\renewcommand{\arraystretch}{1.06}
\resizebox{\columnwidth}{!}{%
\begin{tabular}{l|cc|ccc}
\toprule
Configuration & Optim Params (M) & Total Params (M) & PDMS$\uparrow$ & EPDMS$\uparrow$ & HD-Score Avg.$\uparrow$ \\
\midrule
DrivoR & 18.8 & 41.8 & 93.1 & 48.3 & 35.7 \\
B-II & 17.2 & 39.9 & 91.9 & 45.6 & 33.8 \\
RiSE-Drive & \textbf{18.9} & \textbf{42.3} & \textbf{94.2} & \textbf{51.3} & \textbf{39.5} \\
\bottomrule
\end{tabular}}
\caption{Parameter-efficiency of RiSE-Drive relative to DrivoR and the reduced baseline.}
\label{tab:ablation_params_single}
\end{table}

\section{Experiment}

We build on the DrivoR camera-only pipeline and keep its perception backbone and proposal generator unchanged. Unless otherwise stated, we use four surround-view RGB cameras, a DINOv2 ViT-S backbone~\cite{ref:dinov2} with 16 camera-aware register tokens per camera, 4-layer transformer decoders with hidden dimension 256, and LoRA~\cite{ref:lora} of rank 32 following~\cite{ref:robustbirdseyeview}. For RiSE-Drive, we set $K=64$, $L=4$, $M=10$, $d_{\omega}=d_c=d_p=d_h=256$, $\alpha=0.2$, $\gamma=0.99$, $\lambda_{\mathrm{tail}}=0.5$, $\lambda_{\mathrm{var}}=0.1$, $\delta=0.1$, and $\lambda_b=\lambda_q=\lambda_r=1.0$. The model is trained with AdamW at learning rate $2\times10^{-4}$ using cosine decay on 4 RTX 4090 GPUs.

\subsection{Benchmarks}

\noindent\textbf{NAVSIM-v1.}
We conduct our main evaluation on NAVSIM-v1~\cite{ref:Human-driver}, a large-scale end-to-end planning benchmark instantiated on OpenScene~\cite{ref:OpenScene}, which redistributes annotated driving data from nuPlan~\cite{ref:nuPlan}. Unlike pure trajectory-matching evaluation, NAVSIM-v1 executes the predicted trajectory in a short non-reactive rollout and scores the resulting behavior, directly assessing safety, drivability, smoothness, and route progress. We report the \textbf{Predictive Driver Model Score (PDMS)} together with its main sub-metrics: \textbf{No at-fault Collision (NC)}, \textbf{Drivable Area Compliance (DAC)}, \textbf{Time-to-Collision (TTC)}, \textbf{Comfort (Comf.)}, and \textbf{Ego Progress (EP)}; EP is normalized by a safe upper bound estimated by PDM-Closed~\cite{ref:PDM-Closed}.

\noindent\textbf{NavSafe Split.}
Beyond the standard \texttt{navtest} split, we also evaluate on \texttt{navsafe}, a diagnostic benchmark proposed by Centaur~\cite{ref:Centaur}. It is built by mining difficult cases from \texttt{navtest}, aligning them with NHTSA pre-crash categories~\cite{ref:NHTSA-precrash}, and manually verifying the selected clips, thereby emphasizing rare safety-critical scenarios such as roundabouts, unprotected left turns, ramps, yellow-light decisions, lane-less areas, rare traffic signs, adverse weather, overtaking, and yielding. Compared with aggregate \texttt{navtest} scores, \texttt{navsafe} offers a more sensitive probe of long-tail robustness.

\noindent\textbf{NAVSIM-v2.}
We further evaluate on NAVSIM-v2~\cite{ref:NAVSIM-v2}, which extends NAVSIM-v1 from short-horizon non-reactive scoring to a pseudo-simulation setting. Starting from real logged observations, it synthesizes future sensor views around plausible ego states via neural rendering, exposing planners to perturbed observations that mimic off-distribution rollouts. Its main metric, \textbf{Extended Predictive Driver Model Score (EPDMS)}, augments \textbf{NC}, \textbf{DAC}, \textbf{TTC}, and \textbf{EP} with \textbf{Driving Direction Compliance (DDC)}, \textbf{Traffic Light Compliance (TLC)}, \textbf{Lane Keeping (LK)}, \textbf{History Comfort (HC)}, and \textbf{Extended Comfort (EC)}. On the real-only \texttt{navtest} setting, following prior work, we report the Stage-I score \(\mathrm{EPDMS}_{1}\).

\noindent\textbf{NavHard-Two-Stage Split.}
We also report results on navhard-two-stage, the official hard split of NAVSIM-v2~\cite{ref:NAVSIM-v2}. This split targets interaction-heavy and recovery-demanding cases with a two-stage protocol: in \textbf{Stage I}, the planner is evaluated on the real observation and produces a 4-second endpoint; in \textbf{Stage II}, it is re-evaluated on synthetic branches sampled around plausible Stage-I futures, and the resulting outcomes are aggregated by proximity to the Stage-I endpoint to form the final \textbf{EPDMS}. It is therefore more discriminative than real-only evaluation for measuring robustness under state perturbations.

\noindent\textbf{HUGSIM.}
To complement the NAVSIM family, we additionally evaluate on HUGSIM~\cite{ref:HUGSIM}, a photo-realistic closed-loop benchmark built from reconstructed real-world driving scenes. HUGSIM lifts multi-view RGB observations into interactive 3D environments via Gaussian-splatting-based scene reconstruction and supports iterative policy execution, so the planner's actions affect subsequent observations. The benchmark contains 70+ reconstructed scenes and 400+ scenarios derived from KITTI-360~\cite{liao2022kitti}, Waymo~\cite{sun2020scalability}, nuScenes~\cite{caesar2020nuscenes}, and PandaSet~\cite{xiao2021pandaset}, and is divided into \textbf{easy}, \textbf{medium}, \textbf{hard}, and \textbf{extreme} difficulty levels. We report both \textbf{Road Completion (RC)} and the \textbf{HUGSIM Driving Score (HD-Score)}, where HD-Score summarizes no-collision, drivable-area compliance, TTC, comfort, and global route completion. Compared with NAVSIM-style evaluation, HUGSIM more directly measures long-horizon stability, recovery ability, and robustness to compounding closed-loop errors.

\subsection{Comparison with State-of-the-Art}

Tables~\ref{tab:navsim-v1}--\ref{tab:rc-hdscore-breakdown} compare RiSE-Drive with recent end-to-end planners on NAVSIM-v1~\cite{ref:Human-driver}, NAVSIM-v2 navhard split~\cite{ref:NAVSIM-v2}, and HUGSIM~\cite{ref:HUGSIM}. The compared methods span lightweight CNN backbones, stronger vision encoders such as InternImage-Tiny~\cite{ref:InternImage}, V2-99~\cite{ref:V2-99}, and DINOv2 ViT-S~\cite{ref:dinov2}, as well as larger VLM/VLA backbones such as Emu3-8B~\cite{ref:Emu3} and InternVL3-2B~\cite{ref:InternVL3}. RiSE-Drive uses the same backbone and proposal generator as DrivoR~\cite{ref:DrivoR}; the comparison against DrivoR therefore isolates the effect of long-horizon, distributional, and risk-sensitive candidate evaluation.

\paragraph{NAVSIM-v1.}
Table~\ref{tab:navsim-v1} reports results on the \texttt{navtest} split. RiSE-Drive achieves the best PDMS of 94.2, improving over DrivoR by 1.1 points. Under the matched ViT-S setup, it raises NC from 98.9 to 99.5 and TTC from 96.2 to 98.4 while maintaining perfect comfort; EP decreases slightly from 89.1 to 88.7, indicating a safer and slightly more conservative selection policy.

\paragraph{NAVSIM-v2 \texttt{navhard-two-stage}.}
Table~\ref{tab:navsim-v2-navhard-two-stage} shows that RiSE-Drive achieves the best EPDMS of 51.3, outperforming DrivoR by 3.0 points and the strongest previous result in the table by 3.2 points. The gain is concentrated on Stage-II safety terms, including NC (90.2$\rightarrow$91.5), DDC (91.9$\rightarrow$94.6), and TTC (88.0$\rightarrow$91.3), while the modest EP reduction (70.0$\rightarrow$68.7) again reflects more cautious decisions under perturbation.

\paragraph{HUGSIM.}
Table~\ref{tab:rc-hdscore-breakdown} evaluates fully closed-loop transfer on HUGSIM. RiSE-Drive achieves the best average RC and HD-Score of 50.1 and 39.5, exceeding DrivoR by 0.3 and 3.8 points, respectively. The gain is largest on the hard and extreme subsets, where HD-Score rises from 18.8 to 22.7 and from 32.5 to 39.2, indicating stronger long-horizon robustness under compounding errors.

Overall, the three benchmarks consistently show that, once a strong proposal set is available, improving candidate evaluation yields larger gains than backbone scaling alone.

\subsection{Ablation Study}

Unless otherwise stated, all ablations keep the DrivoR encoder and proposal generator fixed, so the results isolate the contribution of the proposed evaluator. We study three aspects: component-wise contribution, parameter overhead, and sensitivity to the future-branch and quantile design. We denote the original DrivoR by \textbf{B-I}, and a reduced baseline that replaces its scoring decoder with a lightweight direct score head by \textbf{B-II}. Starting from B-II, we progressively add \textbf{FCG} (Future-Condition Generator), \textbf{CTE} (candidate trajectory-level encoder), \textbf{SLE} (step-level encoder), \textbf{LCP} (long-horizon consequence propagation), \textbf{DOE} (distributional outcome estimation), and \textbf{RSR} (risk-sensitive ranking).

\paragraph{Component-wise ablation.}
Table~\ref{tab:ablation_main_single} shows steady gains as the evaluator is built up from the reduced baseline. Future conditioning and trajectory encoding recover much of the gap introduced by simplifying the original scorer, while explicit long-horizon consequence propagation yields the clearest benefit on the harder NAVSIM-v2 setting. The final improvements from DOE and RSR show that distributional prediction and risk-sensitive ranking are complementary rather than interchangeable.

\paragraph{Parameter overhead.}
Table~\ref{tab:ablation_params_single} shows that RiSE-Drive improves all three aggregate benchmarks with only a marginal increase in parameter count over DrivoR. The gain is therefore structural, not a consequence of brute-force model scaling.

\subsection{Visualization}

We further visualize RiSE-Drive on four representative safety-critical cases from NavSafe~\cite{ref:Centaur}: \textbf{UNPL} (\emph{unprotected left turn}), \textbf{EXR} (\emph{exit ramp}), \textbf{YLD} (\emph{yielding}), and \textbf{RDBT} (\emph{roundabout}). As shown in Fig.~\ref{fig:rise_navsafe_vis}, the four columns from left to right correspond to (A) UNPL, (B) EXR, (C) YLD, and (D) RDBT. Across these interaction-heavy long-tail scenarios, RiSE-Drive remains smooth, route-consistent, and conservative in conflict regions. In UNPL and RDBT, it commits only after sufficient clearance to surrounding traffic; in EXR, it follows the ramp geometry stably; in YLD, it resolves right-of-way cautiously before resuming progress. In all four cases, the planned trajectory stays close to the human future trajectory while preserving adequate safety margins, qualitatively supporting the gains on safety-critical scenarios.

\section{Conclusion}

We presented RiSE-Drive, a risk-sensitive evaluator for candidate-based end-to-end autonomous driving. The central premise of this work is that, once a planner already provides a strong set of feasible trajectory proposals, the remaining bottleneck is no longer only candidate coverage, but how those candidates are judged under long-horizon uncertainty. Existing selectors largely compress candidate quality into deterministic point scores or heuristic sub-scores, which can obscure delayed failures and hidden lower-tail risk. RiSE-Drive addresses this limitation by modeling candidate quality as a distribution over future-condition-dependent long-horizon outcomes. Concretely, it predicts a compact set of future-condition branches, performs candidate-conditioned latent consequence propagation, estimates ordered outcome quantiles, and ranks trajectories with a risk-sensitive criterion that combines the mean, lower-tail CVaR, and cross-condition instability.

This formulation makes candidate selection explicitly sensitive to tail failures rather than only nominal performance. As a result, RiSE-Drive prefers trajectories that remain reliable not only on average, but also under plausible downstream deviations and interaction-heavy rollouts. Across NAVSIM-v1, NAVSIM-v2, and HUGSIM, as well as qualitative analysis on safety-critical NavSafe scenarios, RiSE-Drive consistently improves robustness under a matched DrivoR setup, with especially clear gains on hard, perturbation-heavy, and closed-loop evaluations. These results suggest that robust end-to-end driving depends not only on stronger perception backbones or larger models, but also on principled long-horizon candidate evaluation. We hope this perspective encourages future work on tighter integration between proposal generation, consequence modeling, and risk-sensitive decision making.

\bibliographystyle{ACM-Reference-Format}
\bibliography{main} %

\label{LastMainPaperPage}
\clearpage
\appendix
\input{appendix}

\end{document}

%% file: appendix.tex
\setcounter{table}{0}
\renewcommand{\thetable}{A.\arabic{table}}
\renewcommand{\theHtable}{A.\arabic{table}}
\setlength{\parindent}{0pt}
\setlength{\parskip}{5pt}
\fancypagestyle{risesupplement}{%
  \fancyhf{}%
  \fancyhead[L]{\sffamily\fontsize{6.974}{8}\selectfont RiSE-Drive: Risk-Sensitive Long-Horizon Candidate Evaluation for End-to-End Autonomous Driving}%
  \fancyhead[R]{\sffamily\fontsize{6.974}{8}\selectfont Supplementary Material}%
  \renewcommand{\headrulewidth}{0pt}%
  \renewcommand{\footrulewidth}{0pt}%
}
\pagestyle{risesupplement}
\thispagestyle{risesupplement}

\section{Supplementary Material}
\label{sec:supplementary}

\subsection{HUGSIM Results}
Table~\ref{tab:rc-hdscore-breakdown} corrects the HUGSIM Easy RC for RiSE-Drive, which was misprinted as 79.5 in the conference version.

\subsection{Future-Condition Branch Construction}
RiSE-Drive uses $L=4$ reference conditions to construct branch targets (Section~4.6). We pair two motion measurements from the same training-log window. Longitudinal pressure is the negative of the lead vehicle's longitudinal acceleration. Lateral approach is the lateral velocity of a same-direction adjacent vehicle, signed positive toward the ego corridor.

We first bisect the windows by the empirical rank of longitudinal pressure, then bisect each half by the within-group rank of lateral approach. The four conditions in Table~\ref{tab:joint_traffic_conditions} use low and high levels relative to the training logs, with a separate lateral threshold in each longitudinal group.

For each condition, we choose the observed pair closest to the group's componentwise median, measuring distance after scaling both quantities by their training-set interquartile ranges.

\newpage
\begin{table}[t]
\captionsetup{labelsep=period,justification=raggedright,singlelinecheck=false}
\caption{Four joint traffic conditions.}
\label{tab:joint_traffic_conditions}
\centering
\fontsize{8.3}{10}\selectfont
\setlength{\tabcolsep}{2pt}
\renewcommand{\arraystretch}{1.05}
\begin{tabular*}{\linewidth}{@{}c@{\extracolsep{\fill}}lcc@{}}
\toprule
Branch $l$ & Condition & \shortstack{Longitudinal\\pressure} & \shortstack{Lateral\\approach} \\
\midrule
1 & C0 Open passage & Low & Low \\
2 & C1 Longitudinal constraint & High & Low \\
3 & C2 Lateral approach & Low & High \\
4 & C3 Combined constraint & High & High \\
\bottomrule
\end{tabular*}
\end{table}

Each representative sets the lead vehicle's target longitudinal acceleration and the adjacent vehicle's target lateral velocity. Starting from a shared initial state, the benchmark applies these targets smoothly under its road and dynamics constraints. An axis is active only when the relevant vehicle and a feasible motion are available.

Every candidate $\tau_i$ is evaluated under the same four conditions. Benchmark rollout and scoring produce one target $y^{\star}_{i,l}$ per candidate and branch. The targets supervise the corresponding predictions $\hat{y}_{i,l}$ and form an equally weighted empirical outcome distribution for each candidate. At inference, the learned branch code $\omega_l=P_{\mathrm{ctx}}^{(l)}(\bar{\mathbf r}_t)$ is computed from the current scene (Section~4.2).